\documentclass[conference]{IEEEtran}
\IEEEoverridecommandlockouts
\usepackage{cite}
\usepackage{amsmath,amssymb,amsfonts}
\usepackage{algorithmic}
\usepackage{graphicx}
\usepackage{textcomp}
\usepackage{caption}
\usepackage{subcaption}
\usepackage{xcolor}
\usepackage{url}
\def\BibTeX{{\rm B\kern-.05em{\sc i\kern-.025em b}\kern-.08em
    T\kern-.1667em\lower.7ex\hbox{E}\kern-.125emX}}
\begin{document}

\title{A web-based gamification of upper extremity robotic rehabilitation\\
% \thanks{Identify applicable funding agency here. If none, delete this.}
}

\author{\IEEEauthorblockN{1\textsuperscript{st} Payman Sharafianardakani}
\IEEEauthorblockA{\textit{School of Electrical and Computer Engineering, College of Engineering} \\
\textit{University of Tehran}\\
Tehran, Iran \\
paymansharafian@gmail.com}

\IEEEauthorblockN{2\textsuperscript{nd} Hadi Moradi, Fariba Bahrami}
\IEEEauthorblockA{\textit{School of Electrical and Computer Engineering, College of Engineering} \\
\textit{University of Tehran}\\
Tehran, Iran \\
 moradih@ut.ac.ir, fbahrami@ut.ac.ir}
\and
\IEEEauthorblockN{3\textsuperscript{nd} Malahat Akbarfahimi}
\IEEEauthorblockA{\textit{School of Rehabilitation Sciences
Iran} \\
\textit{University of Medical Sciences}\\
Tehran, Iran \\
malahatfahimi@gmail.com}

}

\maketitle

\begin{abstract}
In recent years, gamification has become very popular for rehabilitating different cognitive and motor problems. It has been shown that rehabilitation is effective when it starts early enough and it is intensive and repetitive. However, the success of rehabilitation depends also on the motivation and perseverance of patients during treatment. Adding serious games to the rehabilitation procedure will help the patients to overcome the monotonicity of the treatment procedure. On the other hand, if a variety of games can be used with a robotic rehabilitation system, it will help to define tasks with different levels of difficulty with greater variety. In this paper, we introduce a procedure for connecting a rehabilitation robot to several web-based games. In other words, an interface is designed that connects the robot to a computer through a USB port. To validate the usefulness of the proposed approach, a researcher-designed survey was used to get feedback from several users. The results demonstrate that having several games besides rehabilitation makes the procedure of rehabilitation entertaining.
\end{abstract}


\section{Introduction}
Stroke or Cerebrovascular was the second leading cause of mortality and disability globally until 2012. Over 40\% of stroke survivors encounter medium to critical disorders that need intensive care, and 10\% of them require long-term care in nursing homes or other facilities \cite{ref1}. After a stroke, Survivors enter the rehabilitation process. The purpose of standard rehabilitation is to help patients till recovering their lost ability promptly. A variety of rehabilitation techniques are utilized to recover patients' abilities, and one of them is robotic rehab. Therefore, there are many studies investigating rehabilitation robots ranging from exoskeletons to end-effector robots. The choice of a robot or robotic technology depends on the type of disability.

Today, despite many advances in rehabilitation, many rehabilitation centers use traditional and time-consuming methods. In addition to forcing physiotherapists to monitor and intervene constantly, these methods also make patients tired and reluctant to perform the assigned tasks. Thus, these tasks seem meaningless and monotonous to patients. This reluctance causes patients not to make good progress in recovering quickly. Due to the importance of this issue, one of the proposed methods in rehabilitation or training is gamification. 

Using game mechanics and playful thinking with aesthetics to motivate, learn, and solve problems is called gamification \cite{ref2}. In other words, transforming an action that has no conceptual connection with a game into an action associated with the game is called gamification. For instance, gamification was made in a study on the stairs of a metro station in Stockholm, Sweden. Each step was arranged to the sound of a musical note. When people stepped on the steps, they played the corresponding note. In the end, the results showed that 66\% more people used music stairs instead of escalators \cite{ref3}. Serious games are very similar to gamification, and you cannot find a boundary between them. However, in general, serious games are full-blown games that offer education, not just entertainment.

In this study, we focused to gamify upper extremity rehabilitation using a robotic arm and available games on the web. The main advantage of our proposed approach to the current systems is in using available games on the web by connecting the robotic arm through a USB port to computers. This approach reduces the cost of game development and increases the available options to patients and therapists to fit every need.

\section{Related work}
For many years gamification has been used in education, and results of that have been demonstrated. The effects of gamification on students have shown that they are more enthusiastic and motivated than before \cite{ref2}. These results cause to use of this method in other fields. One of these fields is the field of rehabilitation. The use of gamification in rehabilitation robotics has abundantly increased and we can see vestigial of gamification even in commercialized robots. Robots such as Armeo Power and Armeo Spring from Hocoma company \cite{ref4}, Inmotion ARM \cite{ref5}, and Robotherapist 2D \cite{ref6}, with the help of games, encourage patients to perform rehabilitation movements. These games motivate and help the treatment process to be noun-uniform. Housman et al. \cite{ref7}, surveyed a group of patients and demonstrated that 90\% of patients agree with game-based robotic rehabilitation. Furthermore, LaPiana et al. \cite{ref8} showed that more than 90\% of patients enjoy and engage with games in the rehabilitation process. Also, Nguyen et al. \cite{ref9} evaluated a rehabilitation process with 22 subjects in 2 methods: with gamification and without gamification. In the end, the result indicates a more favorable consequence on the gamified application. Nevertheless, there was no statistical difference in user engagement scores.

Gamification methods are diverse but generally, they can be categorized in three methods: 1)Virtual reality (VR), 2) Augmented Reality (AR) 3) Common gamification. 
\subsubsection{Virtual reality}
VR is a simulation of a world apart from reality that allows the user to enter a space different from his surroundings and have the desired experience; An experience that can be fun and entertaining with the purpose of scientific, educational, healthcare, and music. VR can be an advanced version of human-computer interaction in which users are immersed in a virtual 3D environment containing digital objects \cite{ref10}. In VR, sometimes it is hard to differentiate between the real and unreal world, and several devices such as earphones and head-mounted shows (HMIDs) help to improve this experience. In the field of stroke rehabilitation, VR has many benefits. For example, Dias et al. \cite{ref11} demonstrated that using VR helps people by increasing their motivation for recovery through the use of a fun and comfortable environment. VR encourages patients to maintain the therapy process and not lose motivation in the long run. VR games are designed in such a way that patients are constantly challenged and immersed to attempt harder. Although VR games have many benefits, one of their main disadvantages is their expensive equipment. Furthermore, some cases have also observed that the elderly do not show interest in this method \cite{ref12}. That is why we did not select this method since our target group in this paper is older people. 
\subsubsection{Augmented reality} AR is a combination of virtual reality and physical reality or physical objects.  In AR, the user clearly sees the difference between the virtual and real worlds and has more control over virtual objects by communicating with real objects. AR also plays an essential role in the rehabilitation of stroke patients. For example, Khademi et al. \cite{ref13} utilized a haptic method with AR to demonstrate improved hand stiffness. On the other hand, Mousavi et al. \cite{ref14} experimented with the two groups. The first group used AR and the second group used a combination of mouse and computer. They eventually showed that motor movements increased more in the AR group than in the mouse and computer-based method. Although AR games have many benefits, the most popular augmented reality headsets on the market today are bulky, expensive, and require a computer connection, making the whole process limited and inconvenient for people. Consumers may also utilize AR software on their smartphones or tablets. However, showing graphics correctly in mobile AR is a huge challenge. Electric interference, for example, can disrupt mobile sensors such as accelerometers, which is frequent in metropolitan settings. Furthermore, augmented reality apps may be extremely distracting and can cause physical harm. Many individuals have been wounded while playing Pokemon Go, for example \cite{ref15}. Similarly, augmented reality apps might result in significant casualties when employed in potentially dangerous locations such as busy roadways, construction sites, and medical facilities.
\subsubsection{Common Gamification} CG is a method that is usually used on orthotic robots like exoskeletons and end-effectors. In this method, the robots connect to a computer, and the games simulate their movements. For instance, Inmotion Arm \cite{ref5}, developed by Bionic and MIT, uses the CG method. Inmotion Arm is one of the most successful examples in recent years and is widely used in 20 countries worldwide \cite{ref5}. In addition to having several games to motivate patients, the robot also has a robust evaluation system that records patient movement feedback. One of these games is Clock \cite{ref16}, which is very simple and helpful. The practical results of this robot worldwide have shown that using the robot along with the CG method will improve patients in the long run. On the other hand, Instead-Technology in Robotherapist 2D did the CG method, And they used some games to motivate patients. They also demonstrated that this using Robotherapist 2D besides these games help patients to improve their upper-extremity motor activity \cite{ref16}.

Examination of different methods shows us that each method has its own essential features, and according to the type of rehabilitation, one of these methods should be selected for gamification. However, it can be said that the VR method and the CG method are more appropriate for the exoskeleton and end-effector rehabilitation robots. However, the VR method is more expensive, while we always have access to the desired equipment in the CG method. So in this investigation, we finally chose the CG method.

As it was mentioned earlier, the first and foremost purpose of gamification is to increase patients' motivation and therapy effectiveness \cite{ref17,ref18,ref19}. Using games in the rehabilitation process makes rehabilitation tasks meaningful. On the other hand, having a game to continue rehabilitation causes the treatment process to be monotonous. So having multiple games is very critical. The method presented in this article helps us access many games and ultimately increases the variety in rehabilitation. However, the main advantage of our approach is using simple available games. Furthermore, we chose the CG method because we need a low-cost device, thus this method has low cost and helps therapists have an effective rehabilitation. In this regard, therapists can choose between the available games according to the patient's ability and give the patient the right to choose between several games. As a result, a therapist has many options for effective treatment, and patients are satisfied with the treatment process.

\section{Web-based games}
As mentioned earlier, gamification as an effective method increases motivation and makes the rehabilitation process enjoyable. One of the challenges in the gamification of rehabilitation robots is the number of games used. Using a few games makes the rehabilitation process monotonous and tedious for patients in the long term. Our method to solve this challenge is to use web games. Web games are actually a kind of web service. Web services allude to a supplier that gives the ability to perform particular operations through a browser. Consumers may reach services at any time and place through the internet and web services and access locally configurable online features \cite{ref1}. Web services are divided into two kinds websites and web applications. A web application is an application that can be accessed via the Internet anywhere, anytime. The main concept of a web application is to overcome the limitations of local gadgets utilizing remote computing power \cite{ref1}. One of the demos that use remote computing power is a web game. Web games are not similar to online games; while online games have to be downloaded the game and installed on the local computer for the game to play, web games are launched without installation in a web browser \cite{ref1}. These features make web games a good choice for our rehabilitation robot. This article provides a way for a rehabilitation robot with one or two degrees of freedom to connect to web games and organize a web-based rehabilitation framework based on the features of these games.

\begin{figure}[ht]
    \centering
    \includegraphics[width=0.6\linewidth]{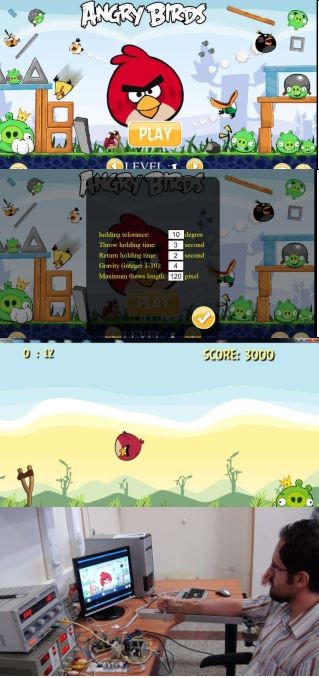}
    \caption{The previous game used on the robot based on open source version of the Angry Birds game}
    \label{fig1}
\end{figure}

\section{Methods}
The method presented in this paper is to design and program the robotic arm that can be connected through the USB port to any computer and control games like a mouse. . Using this method has given us access to a large number of games.  In this way, we used the serial port to have a pure connection between the computer and the robot. One of the most significant advantages of this kind of connection is stability. Furthermore, this method is cheap because we have access to computers everywhere we want. Consequently, with the pure connection and many games, we have a comprehensive system that helps us decrease monotonously in rehabilitation.

In the past, this robot worked with a few games that were designed exclusively for it (Fig. \ref{fig1}). There are always two main problems in rehabilitation. 1. get patients used to help 2. uniformity of treatment steps. For the first problem, an active resistance control has been used to challenge patients more. While using the method presented in this study has made the patient take more initiative and think more to overcome the challenges, in the methods provided for other robots, pre-determined goals were always set for patients, and robots always help patients. This is like we are constantly cycling with auxiliary wheels. Therefore, we will never be able to ride a bike independently.  For the second problem, we provide the method that we explained in this paper.

\begin{figure}[h!]
    \centering
    \includegraphics[width=0.9\linewidth]{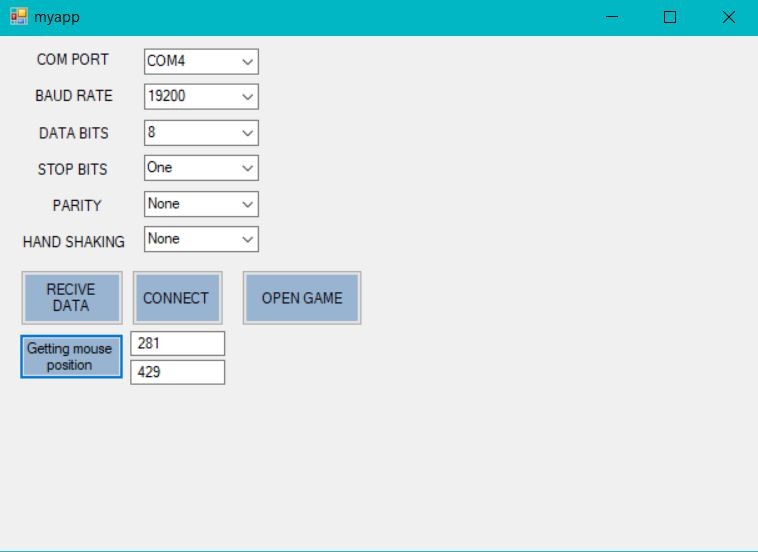}
    \caption{User interface to setup the USB connection between the rehabilitation robotic arm and a computer}
    \label{fig2}
\end{figure}

\subsection{User interface}\label{AA}
For connecting to web games, the first issue is to connect easily. For this purpose, we designed a user interface. We used the C\# Windows form application to have a helpful and straightforward user interface. All parts related to connecting to a computer and storing the data collected from patients are included in this code. As shown in Fig. \ref{fig2}, after connecting the robot to the computer via USB, all port characteristics are automatically detected, and this connection is established only by pressing the connect button. Also, by pressing open game, one of the games opens. All these features make it easy to use the robot and work with the game.

\subsection{Robot description }
\begin{figure}[ht]
     \centering
     \begin{subfigure}[a]{0.35\textwidth}
         \includegraphics[width=\textwidth]{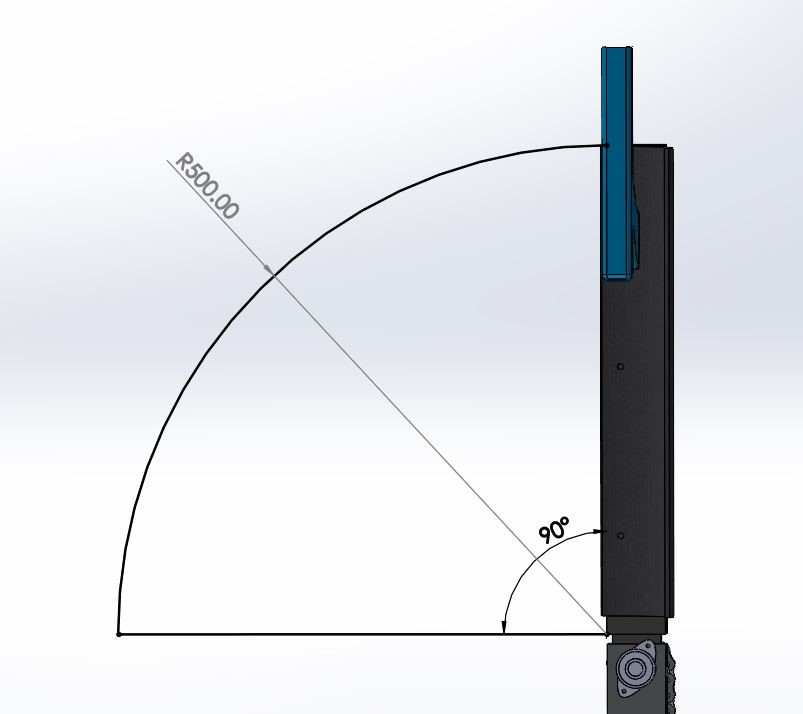}
         \caption{}
         \label{fig3a}
     \end{subfigure}
     \\
     \vspace{0.2cm}
     \begin{subfigure}[b]{0.35\textwidth}
        \centering
        \includegraphics[width=\linewidth]{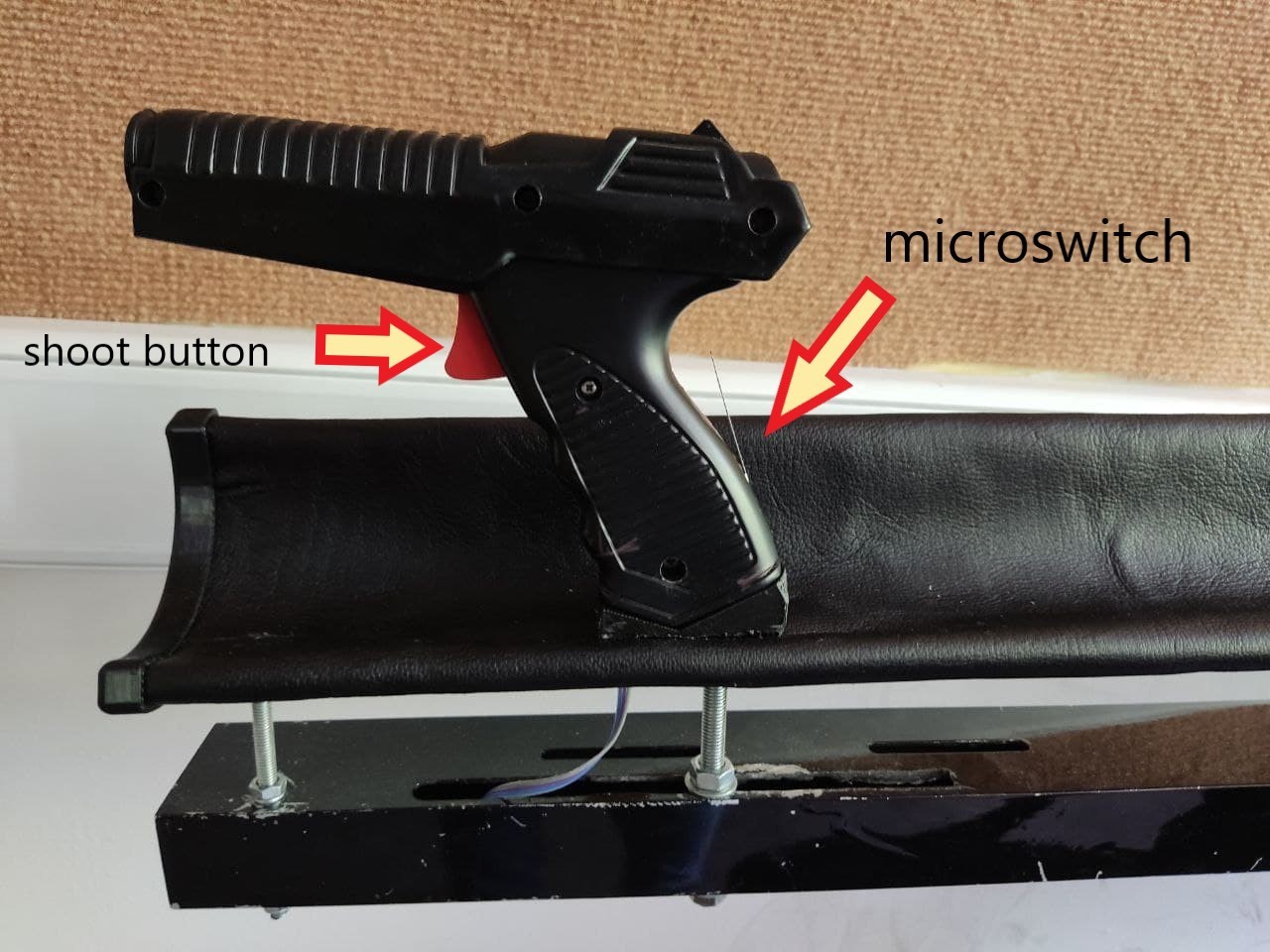}
        \caption{}
        \label{fig3b}
     \end{subfigure}
     \caption{ (a) shows The robot’s workspace, and (b) The joystick including the shoot button and the microswitch }
     \label{fig3}
\end{figure}

The University of Tehran rehabilitation robot is a one-degree-of-freedom robot with two encoders. These encoders show the position of the arm and motor. Our recommended method is to convert the arm position to the mouse's position on the computer screen. In this method, we first positioned the encoders and then mapped the range of the mouse on the screen. Fig. \ref{fig3a} shows the workspace of the robot arm movement, which shows that this movement is limited. Furthermore, to shoot in these games, we turned the button on the joystick to the left click on the computer (Fig. \ref{fig3b}) Additionally, there is a microswitch for safety. When the patients remove their hands from the joystick, the robot and game will stop.

\vspace{-1pt}
\subsection{Games }
\begin{figure}[t]
     \centering
     \includegraphics[width=0.7\linewidth]{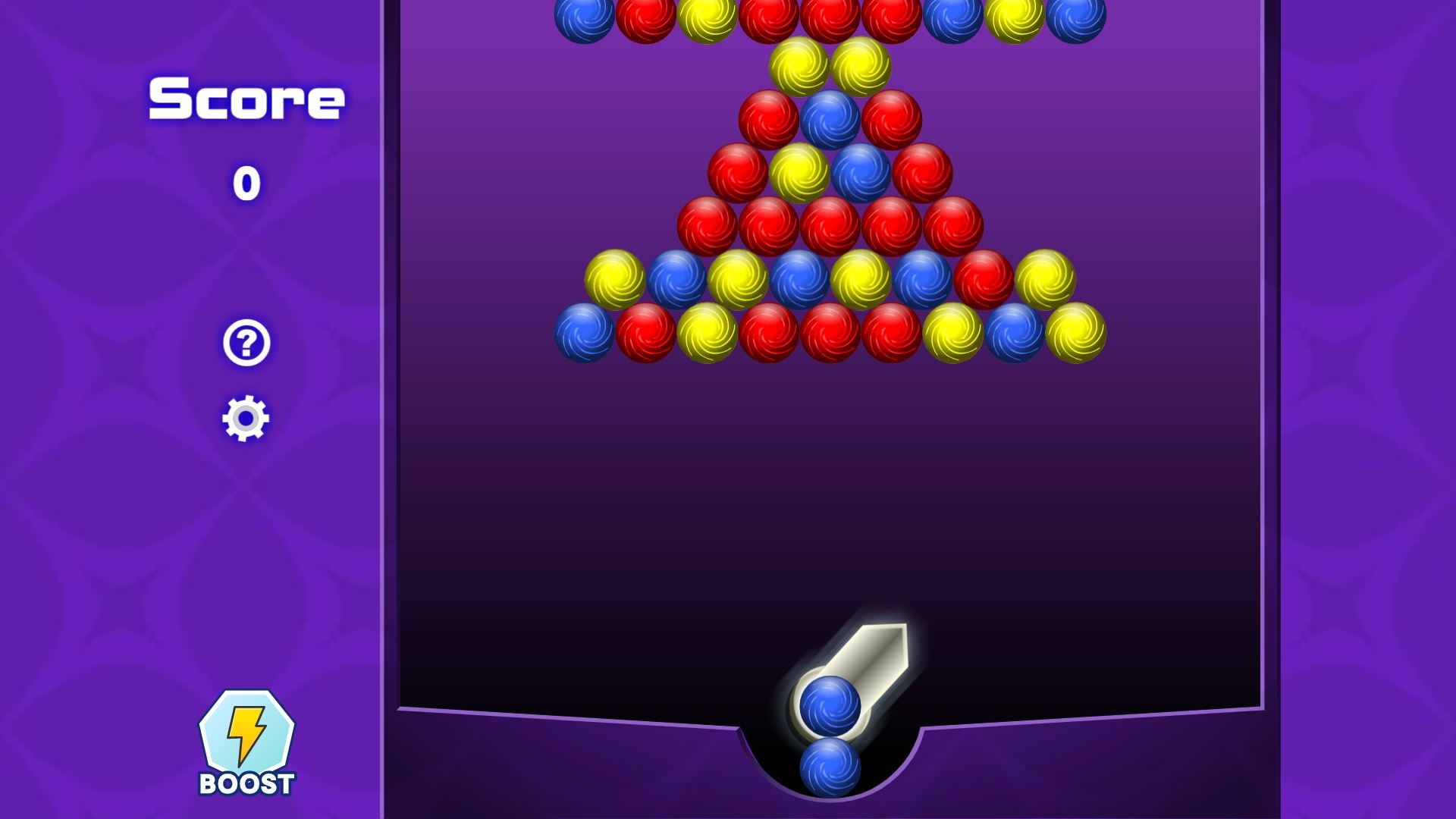}
     \\
     \includegraphics[width=0.7\linewidth]{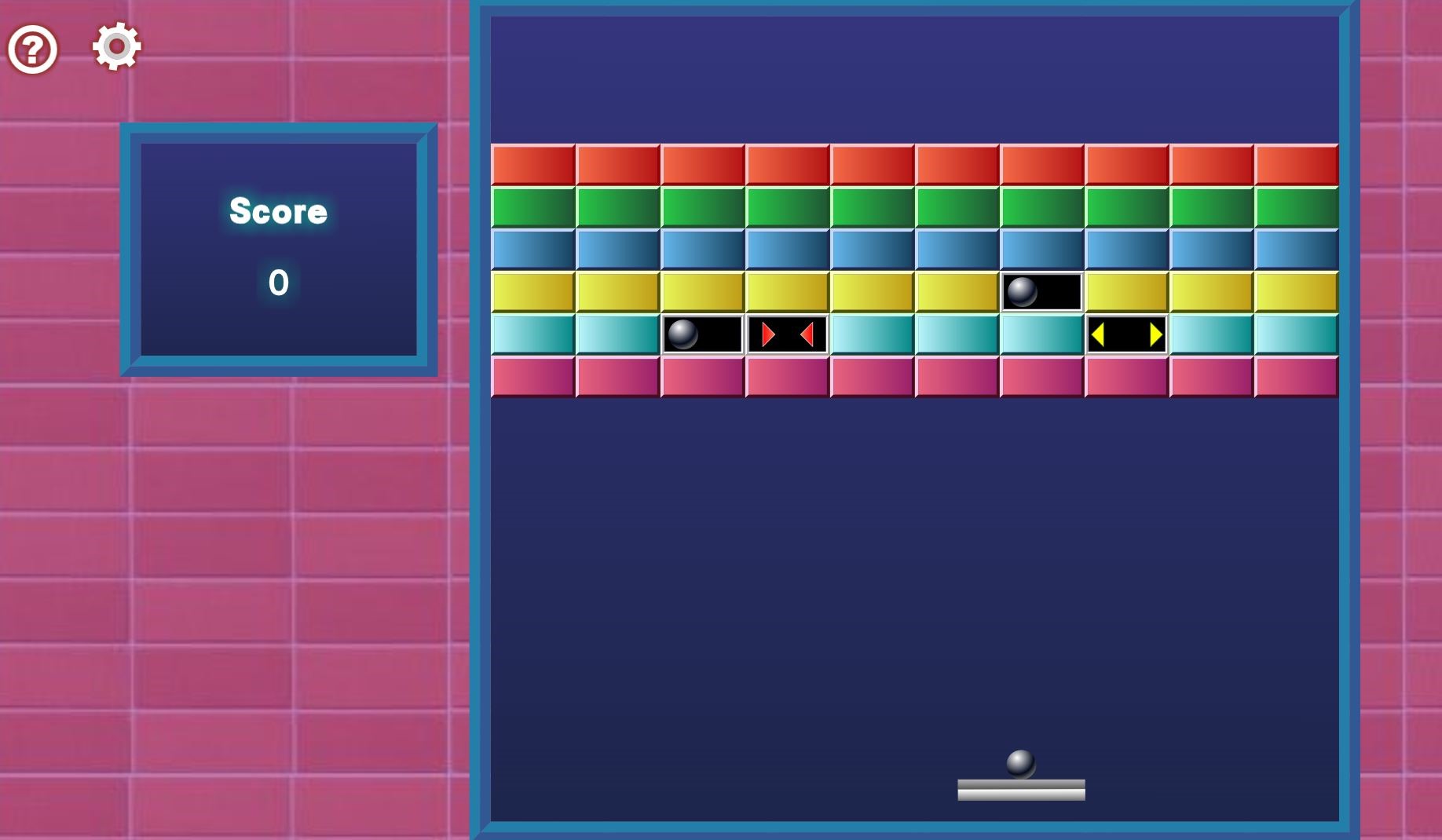}
     \caption{ Two web-based games used in our study connected to our rehabilitation robot \cite{ref20}\cite{ref21}}
     \label{fig4}
\end{figure}

Each of these games has difficulty levels and becomes more difficult after passing each stage. All games are a combination of visual and audio features that excite the user and increase their motivation. Fig. \ref{fig4} illustrates some of these web games \cite{ref20}\cite{ref21}. According to the occupational therapist, these games not only motivate but also engage the user. Because the fact that you always have to choose the best path creates a mental challenge for the user.

\subsection{Experiment and results}
\begin{figure}[hb]
     \centering
     \includegraphics[width=0.9\linewidth]{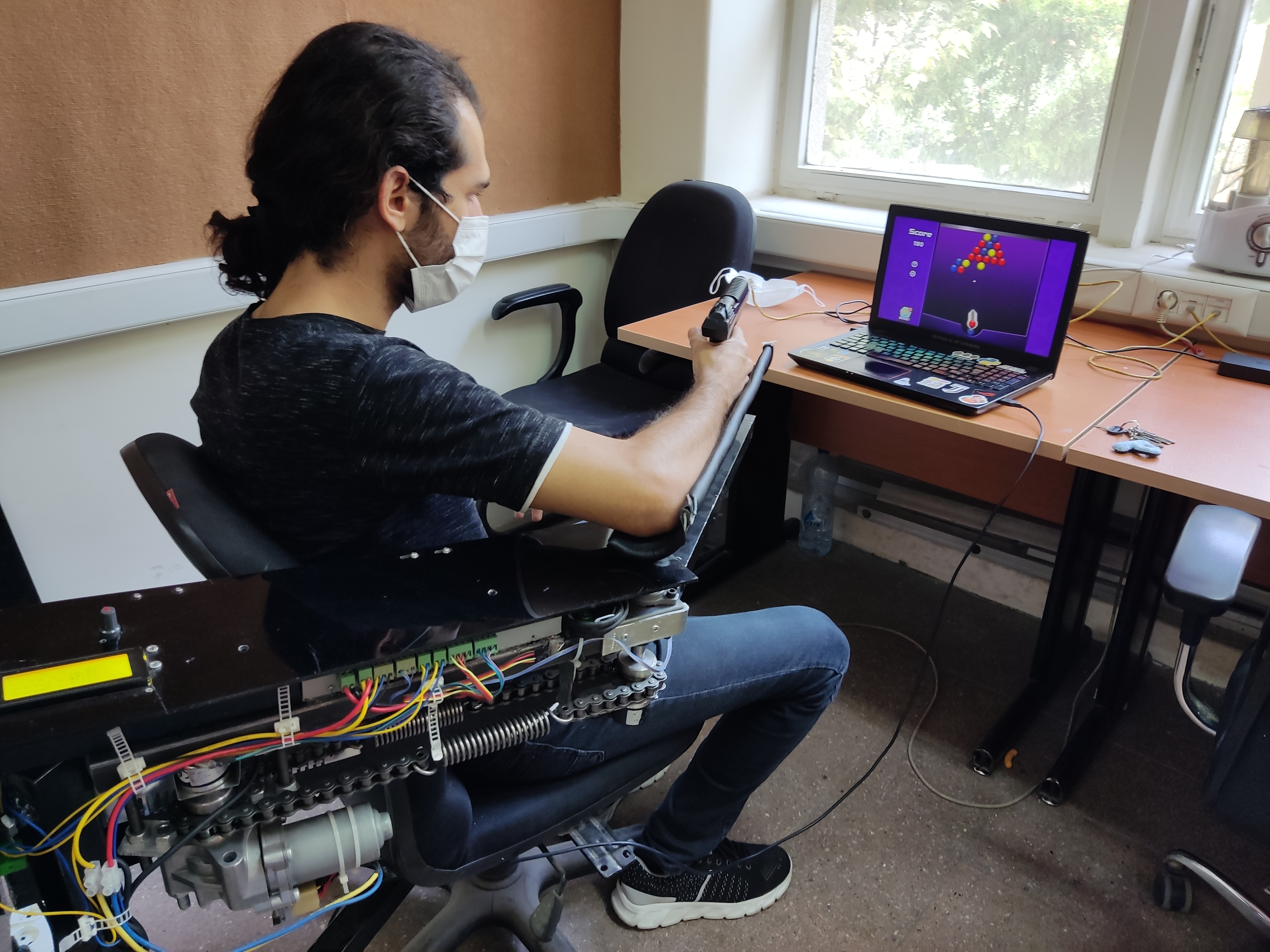}
     \caption{A subject is playing with the robotic arm}
     \label{fig5}
\end{figure}

Finally, we performed an experiment on normal subjects (Fig. \ref{fig5}). We had a protocol, including a survey that subjects filled out at the end of the test.   In this evaluation, we had 15 subjects with the same features. All subjects were in the range of 20 and 30 years old, and they were right-handed. Moreover, they did not have the long antecedent of exercise by hand, and They were not tired or lethargic before the test. Also, this experiment did not have any sexuality limit.
\subsubsection{Protocol}

\begin{table*}[ht]
    \centering
    \begin{tabular}{|p{5cm}|c|c|c|c|c|c|}
        \hline
        \textbf{Type of Questions} & \textbf{Number of Questions} & \textbf{Very Satisfied} & \textbf{Satisfied} & \textbf{Natural} & \textbf{Unsatisfied} & \textbf{Very Unsatisfied} \\
        \hline
        Robot Convenience & 3 & 37.78\% & 42.22\% & 13.33\% & 6.66\% & 0 \\
        \hline
        Robot Safety & 1 & 53.33\% & 33.33\% & 13.33\% & 0 & 0 \\
        \hline
        Making Entertainment and Motivation with Games & 1 & 46.66\% & 53.33\% & 0 & 0 & 0 \\
        \hline
        Concentration & 1 & 53.33\% & 46.66\% & 0 & 0 & 0 \\
        \hline
        Simplicity in Use & 1 & 66.66\% & 33.33\% & 0 & 0 & 0 \\
        \hline
        Tolerance for Tasks and the Degree of Difficulty & 1 & 40\% & 26.66\% & 26.66\% & 6.66\% & 0 \\
        \hline
        Causes Pain and Fatigue & 1 & 33.33\% & 53.33\% & 13.33\% & 0 & 0 \\
        \hline
    \end{tabular}
    \caption{Survey results}
    \label{tab:survey_results}
\end{table*}
The protocol used in this experiment was to perform each person with two games and each game with two different powers. These powers changed after passing two levels in the game, and the whole test period was about 20 to 30 minutes. Because our robot had an active resistance control, we wanted to test both very high and very low power on subjects. As a result, the applied torques were 8 Nm and 16 Nm for each game. Also, during this period, all the movement data of the people were recorded. Finally, after working with the robot, the people filled a questionnaire, and the results of this questionnaire were the average of the answers in each section.

\subsubsection{Results}
In the end, we had a survey with a questionnaire table, which showed that more than 90\% of people were satisfied with multiple games with the robot. This satisfaction results from playing two games for 20 minutes, and the question analyzed the games in terms of entertainment and motivation. Furthermore, in the focus discussion, all subjects believed that they had the necessary focus during the experiment, and some games made them more focused.

According to Table 1, for this questionnaire, we use the Likert scale. We had 9 questions and table 1 shows the subject of every question. In the part making entertainment and motivation, everyone believes that the games provide entertainment and cause to increase motivation. Additionally, in the part of concentration, everyone believes the games provided concentration during the experiment.

\section{Conclusion }
We believe that games can revolutionize rehabilitation systems overall and stroke rehabilitation in specific. These games, along with entertainment, increase patients' motivation. Consequently, patients practice more and recover faster. In this article, we provide a way to connect a large number of games with a robot. As a result, we increased the variety of rehabilitation and allowed the therapist to choose from the available games, whichever is appropriate for the patient's condition. Also, it can offer different games and tasks for the patient on additional days of the week. Finally, in a survey of ordinary people, we measured the accuracy of this gamification. The results of this survey showed that the variety in rehabilitation is satisfactory for many people.


\begin{thebibliography}{99}

\bibitem{ref1}
I. Jung, J. Lee, J. Kim, and C. Choi, "Ubiquitous gamification framework for stroke rehabilitation treatment based on the web service," in Proceedings of the 4th Workshop on ICTs for improving Patients Rehabilitation Research Techniques, Lisbon, Portugal, 2016, pp. Pages.

\bibitem{ref2}
G. Kiryakova, N. Angelova, and L. Yordanova, "Gamification in education," in Gamification in education (Proceedings of 9th International Balkan Education and Science Conference, 2014), pp.

\bibitem{ref3}
W. H.-Y. Huang and D. Soman, "Gamification of education," Report Series: Behavioural Economics in Action, 2013, 29.

\bibitem{ref4}
C. Sicuri, G. Porcellini, and G. Merolla, "Robotics in shoulder rehabilitation," Muscles, ligaments and tendons journal, 2014, vol. 4, pp. 207-213.

\bibitem{ref5}
Y.-W. Hsieh, K.-c. Lin, C. Wu, T.-Y. Shih, M.-w. Li, and C. Chen, "Comparison of proximal versus distal upper-limb robotic rehabilitation on motor performance after stroke: A cluster controlled trial," Scientific Reports, 2018, vol. 8.

\bibitem{ref6}
I. Díaz et al., "Development of a robotic device for post-stroke home tele-rehabilitation," Advances in Mechanical Engineering, 2018, vol. 10, no. 1, p. 168781401775230.

\bibitem{ref7}
S. Housman, K. Scott, and D. Reinkensmeyer, "A Randomized Controlled Trial of Gravity-Supported, Computer-Enhanced Arm Exercise for Individuals With Severe Hemiparesis," Neurorehabilitation and neural repair, 2009, vol. 23, pp. 505-514.

\bibitem{ref8}
N. LaPiana et al., "Acceptability of a mobile phone–based augmented reality game for rehabilitation of patients with upper limb deficits from stroke: Case study," JMIR Rehabilitation and Assistive Technologies, 2020, vol. 7, no. 2.

\bibitem{ref9}
D. Nguyen and G. Meixner, "Comparison User Engagement of Gamified and Non-gamified Augmented Reality Assembly Training," 2020, pp. 142-152.

\bibitem{ref10}
O. Mubin et al., "Exoskeletons with virtual reality, augmented reality, and gamification for stroke patients' rehabilitation: Systematic review," JMIR Rehabilitation and Assistive Technologies, 2019, vol. 6, no. 2.

\bibitem{ref11}
P. Dias et al., "Using Virtual Reality to Increase Motivation in Poststroke Rehabilitation," IEEE Computer Graphics and Applications, 2019, vol. 39, pp. 64-70.

\bibitem{ref12}
S. Syed-Abdul et al., "Virtual reality among the elderly: a usefulness and acceptance study from Taiwan," BMC Geriatrics, 2019, vol. 19, no. 1, pp. 223.

\bibitem{ref13}
M. Khademi et al., "Haptic Augmented Reality to monitor human arm's stiffness in rehabilitation," in Haptic Augmented Reality to monitor human arm's stiffness in rehabilitation, 2012, pp. 892-895.

\bibitem{ref14}
H. Mousavi Hondori et al., "Choice of Human-Computer Interaction Mode in Stroke Rehabilitation," Neurorehabilitation and Neural Repair, 2015, vol. 30, no. 3, pp. 258-265.

\bibitem{ref15}
N. Joshi, "5 challenges to AR adoption, challenges with augmented reality," [Online]. Available: \url{https://www.allerin.com/blog/5-challenges-to-mainstream-ar-adoption}. [Accessed: 28-Nov-2021].

\bibitem{ref16}
F. Badesa et al., "Pneumatic planar rehabilitation robot for post-stroke patients," Biomedical Engineering: Applications, Basis, and Communications, 2014, vol. 26.

\bibitem{ref17}
F. Carneiro et al., "A Gamified Approach for Hand Rehabilitation Device," International Journal of Online Engineering (iJOE), 2018, vol. 14, pp. 179.

\bibitem{ref18}
J. W. Burke et al., "Serious Games for Upper Limb Rehabilitation Following Stroke," in Serious Games for Upper Limb Rehabilitation Following Stroke, 2009, pp. 103-110.

\bibitem{ref19}
D. Rand et al., "Eliciting Upper Extremity Purposeful Movements Using Video Games: A Comparison With Traditional Therapy for Stroke Rehabilitation," Neurorehabilitation and Neural Repair, 2014, vol. 28, no. 8, pp. 733-739.

\bibitem{ref20}
Novel Games, "Bouncing Balls II," [Online]. Available: \url{https://www.novelgames.com/en/bouncing2/}. [Accessed: 28-Nov-2021].

\bibitem{ref21}
Novel Games, "Bricks Squasher," [Online]. Available: \url{https://www.novelgames.com/en/arkanoid/}. [Accessed: 28-Nov-2021].

\end{thebibliography}
\end{document}